\documentclass{article}
\usepackage[utf8]{inputenc}  
\usepackage{fullpage}

\usepackage[export]{adjustbox}
\usepackage{amsfonts}
\usepackage{amsmath}
\usepackage{amssymb}
\usepackage{braket}
\usepackage{color}
\usepackage[greek, english]{babel}
\usepackage{graphicx}
\usepackage{hyperref}
\usepackage[utf8]{inputenc} 
\usepackage{natbib}
\usepackage{setspace}
\newif\ifdraftcolors
\draftcolorstrue

\ifdraftcolors

\else

\fi

\def\surl#1{{\small{\url{#1}}}}

\title{Natural, Artificial, and Human Intelligences}

\author{
  \begin{tabular}{c@{\hskip 2cm}c}
    Emmanuel M. Pothos & Dominic Widdows \\
    \small City, University of London & \small Independent Research \\
    \small \texttt{emmanuel.pothos.1@city.ac.uk} & \small \texttt{dwiddows@gmail.com}
  \end{tabular}
}

\begin{document}
\maketitle



\begin{abstract}
Human achievement, whether in culture, science, or technology, is unparalleled in the known existence. 
This achievement is tied to the enormous communities of knowledge, made possible by language: leaving theological content aside, it is very much true that 
\selectlanguage{greek} ``Ἐν ἀρχῇ ἦν ὁ λόγος'' \selectlanguage{english} (in the beginning was the word),
and that in Western societies, this became particularly identified with the written word. 
There lies the challenge regarding modern age chatbots: they can ‘do’ language apparently as well as ourselves and there is a natural question of whether they can be considered intelligent, in the same way as we are or otherwise. Are humans uniquely intelligent? 
We consider this question in terms of the psychological literature on intelligence, evidence for intelligence in non-human animals, the role of written language in science and technology, progress with artificial intelligence, the history of intelligence testing (for both humans and machines), and the role of embodiment in intelligence. We think that it is increasingly difficult to consider humans uniquely intelligent. There are current limitations in chatbots, e.g., concerning perceptual and social awareness, but much attention is currently devoted to overcoming such limitations.

\end{abstract}

\section*{Keywords}
artificial intelligence, human intelligence, animal intelligence, chatbots, large language models, intelligence testing, Imitation Game, embodiment 

\section{Statement of the Problem: are Chatbots Intelligent?}

We are witnessing one of the most important and transformative technological revolutions in our history, the emergence of artificial intelligence (AI).
AI may be compared with inventions of historic significance such as the printing press, steam engine, and the electronic computer, and even compared with these, the economic and scientific impact of AI is being felt during a much more concentrated period of time.
A discussion with a modern chatbot, such as chat GPT (we refer to large language models as chatbots in a generic way), is undeniably human-like and several studies have been establishing that chatbots display human cognitive traits, including similar biases \citep{binz2023using,suri2024decision}, the recognition of emotional states 
\citep{schaaff2023exploring} and empathy \citep{roshanaei2024talk}, and other mental processing similarities (\cite{mei2024turing,strachan2024testing}). 
One effect of this revolution is existential: we, humans, often assume our place in the world is unique, by virtue of the uniqueness of our intelligence. Is human intelligence unique? 

A more practical question is whether chatbots will take our jobs. Together with being human-like, chatbots also encapsulate, in some sense, all human knowledge. Asking GPT-4 whether it will take my lecturer job, it is humble and reassuring: ``As a university lecturer, you're definitely in a role where expertise, creativity, and human connection are key. While AI, like ChatGPT, can help with things like answering questions, explaining concepts, or generating teaching materials, it's not likely to replace the heart of your job.'' 
However, increasingly in higher education, new policies are introduced to regulate the use of chatbots, given their dual ability to generate plausible coursework and to mark it with some competence (e.g., \citet{gultom2024use}), questioning the role of humans in the process. Moreover, there are already indications that chatbots can carry out research roles \citep{lehr2024chatgpt}. 

The observation that technological advancements can lead to job losses is hardly new --- there is a name for this, ``technological unemployment'' \citep{peters2020beyond}  --- and there are some compelling examples, e.g., the reduction of the U.S. workforce employed in agriculture between 1900 (41\%) and 2000 (2\%, \citet{autor2014skills}).  
Unsurprisingly, such concerns have been raised for chatbots too (Callum, 2016).  The balancing perspective is that technological innovation also leads to new jobs (\cite{autor2015why,mincer2000technology}) and over-productivity could be compensated in other ways (such as universal income). 
Additionally, some jobs require a human touch, do not depend on high levels of intelligence, or are even unnecessary (perhaps the result of vested interests). 
However, there is a unique role of chatbots in this debate in that they are not `just' another technological innovation with potential to eliminate some jobs: in all cases of technological automation, the replacement concerned tasks that were either too laborious (e.g., agricultural tasks) or seen as beyond the competencies which we have considered central to our notion of human intelligence (e.g., multitudes of rote arithmetical operations in accounting). Instead, with chatbots, we are faced with an artificial agent which seems just like us, concerning some competencies which we have considered distinctly human. 

Overall, the consideration of whether chatbots can replace humans in various jobs is important and complements the existential question of whether chatbots are human-like. The present focus is intelligence. Can chatbots be considered intelligent? A difficulty in addressing this question is that our understanding of intelligence is itself incomplete. We consider some relevant ideas below. Briefly, on the one hand, it appears that traditional notions of intelligence can be covered by chatbots -- and, moreover, a careful discussion reveals challenges to a uniqueness premise for human intelligence. On the other hand, the fundamental question is whether there is a special ingredient in human intelligence, which allows us to rise to apparent amazing intellectual accomplishments. What is it about our minds which, for example, has enabled us to invent quantum mechanics? Note, considering (something like) quantum mechanics an impressive intellectual accomplishment is not an anthropocentric bias: quantum mechanics has allowed surprising, transformative insights into the workings of nature, leading to unthinkable inventions such as semiconductors (and so modern computers). We think that the value of such accomplishments is manifest, independently of the fact that it was us who made them.

One possibility might be that there is no `special' ingredient, rather human achievement is gradual and social, 
so we could possibly explain in detail why quantum mechanics and electronic computers would have been almost impossible to invent 
before industrial developments in the 1800s, without assuming that human intelligence is unique as a precondition.

\section{What is Intelligence: Psychological Approaches}

Attempts to define intelligence long predate modern psychology.
In Western philosophy, intelligence has been traditionally connected with reason and thinking \citep[Ch.~1]{gardner2011frames}.
Aristotle associates reasoning particularly with the concept of \selectlanguage{greek}λογος\selectlanguage{english},
a historic connection discussed in Section \ref{ref:western_word}.

For Descartes, awareness of one’s own thoughts proved existence itself (\emph{cogito ergo sum}).
This tradition takes on a modern form in the work of Freud and Jung, who made introspection about the self and its symbolic life central to the emerging science of psychology.

There has also long been a desire for more empirical foundations --- Hume, for example, was particularly
sceptical of self-reported self-awareness as a scientific grounding.
Mathematicians including Leibniz and Boole proposed ways to formalize and even automate reasoning, 
and Whitehead extended such symbolic approaches into broader theories of abstraction and meaning.
These efforts eventually aligned intelligence with symbolic operations that could be both economically advantageous and systematically measured, which we'll see became common practice in the later 1800s.

In psychology, by the late 1800s, figures such as Galton, Wundt, and William James had begun to turn these ambitions into methods, using reaction times, association tests, and other measures to make mental processes experimentally observable.
By this point, Western psychology largely treated intelligence as a property of individual minds, though with different approaches to studying those minds. 

As certain kinds of intelligence could be tested, it became tempting to define intelligence in terms of what such tests could measure.
Much of the psychological tradition regarding intelligence concerns psychometric theories, which aim to explain individual differences in test scores, on the basis of factors (e.g., see \citep{sternberg2012intelligence}). 
The earliest proposal of this kind is due to \citet{spearman1904general}, who argued for two factors, a general one (later associated with `mental energy') and a specific one, comprised of competencies for particular tasks. 
\citet{thurstone1938primary}, instead, identified seven ``primary mental abilities'' as the determinants of intelligence. These mental abilities include verbal fluency, spatial visualization, and perceptual speed. While originally \citet{spearman1904general} and \citet{thurstone1938primary} were interpreted as antithetical, this is not necessarily the case. One way to reconcile the two views is if the general vs. more specific intelligence factors are hierarchical, so that the latter informs specific aspects of the former. This hierarchical approach has been taken up by other intelligence psychologists, such as \citet{carroll1993human}, who expanded both the specific and general abilities.

This literature has recognised that some abilities reflect knowledge, while others processing power (in some general sense). \citet{cattell1943measurement} introduced the distinction between fluid and crystallized abilities, with the former concerning performance that depends on prior knowledge minimally (e.g., problem-solving on a novel task) and conversely for the latter (e.g., vocabulary tasks). There is a converse relation between these two kinds of abilities and age, with fluid decreasing, while crystallized increasing, with age (e.g., \citet{ghisletta2012two}).

A more processing account of intelligence has been that it is characterized by the speed of information processing \citep{deary2001reaction}, 
sometimes in simple perceptual tasks, such as whether two lines are the same length \citep{mackintosh1981measure} (but see \citep{kiesling1988clocking}). 
There have been related proposals; for example, \citet{hunt1973individual} considered the speed with which information is 'absorbed' into memory and subsequently retrieved. 
Such approaches do not depend on extent of knowledge or other culture-specific skills, such as social skills.

There have been other attempts to understand intelligence in more cognitive terms, e.g. in terms of working memory capacity \citep{conway2013working}, underpinned by high correlations between measures of working memory and reasoning ability (e.g., \citep{kyllonen1990reasoning,ackerman2005working}). Interestingly, in cognitive science, working memory capacity \textit{limitations} have been argued to enable learning in complex domains \citep{newport1988constraints}.

Another processing idea is that intelligence relates to the ability to solve problems and achieve goals. \citet{sternberg1985beyond} proposed that intelligence is characterized by three abilities: practical intelligence (environment interaction), creative intelligence (new problems and situations), and analytical intelligence (primarily problem-solving). Problem-solving ability has had a prominent role in other proposals (e.g., \citep{danner2011measuring,hambrick2019problem}), as well as often being a component in most others  (starting from \citep{spearman1904general}). \citet{sternberg1985beyond}'s theory is an example of approaches positing multiple types of intelligence. \citet{gardner2006multiple} proposed a longer list, including musical, interpersonal, spatial, and linguistic intelligence, and \citet{salovey1990emotional} introduced the idea of emotional intelligence as the ability to perceive and utilize emotions in cognition.

Psychometric approaches to intelligence have demonstrated practical value. For example, they predict longevity \citep{deary2001intelligence,deary2007intelligence}, health outcomes \citep{hart2003childhood}, job performance \citep{schmidt2004general,ericsson2016peak}, and academic success \citep{deary2007intelligence}. 

Particular aspects of intelligence have been prioritised in intelligence testing:
the approaches of \citep{spearman1904general}, \cite{binet1905methodes} were translated into educational and cultural settings that emphasized the importance
of written answers to prearranged test questions \citep{rogele2020biographies}. A preference for written demonstrations of intelligence, particularly in
English-speaking educational systems, has profoundly affected the way artificial intelligence has been viewed and pursued, a relationship that we explore throughout the rest of this paper.

Summing up, this is a complex and rich literature, and we only want to touch upon a handful of points, as relevant to this discussion. First, it is assumed that there are cognitive abilities that correlate with intelligence/the intelligences \citep{corr2024personality}. Second, some of these abilities concern processing characteristics, for example, processing speed or memory efficiency. 
Third, intelligence involves several aspects, e.g., problem-solving, linguistic/verbal, emotional, social, etc. Can these insights help us understand what, if anything, sets apart human intelligence compared to biological species and artificial agents? We can eliminate some possibilities. Artificial agents, including chatbots, exceed human capacity in essentially all processing characteristics (speed of processing, memory, information absorption and retrieval, problem solving ability, at least in specific categories), that is, it is unlikely that we can understand human vs. chatbot intelligence in terms of processing characteristics. Regarding types of intelligence, note first that non-human animals often demonstrate the competencies (at least to a degree) cited for humans too, such as spatial, emotional, musical abilities etc. (cf. \citet{gardner2006multiple}). 

Also, work such as Gardner's argues for many types of intelligence, over and above linguistic and logical intelligence. There is certainly a question of whether there are types of human intelligence less possible for chatbots. As we shall see later, psychometric testing of human intelligence has focussed mostly on linguistic and logical intelligence, leading to challenges which are easily won by chatbots, both because of content and format (e.g., multiple choice tests, while cheap because of automated marking, are particularly easy for chatbots). Indeed, regarding psychometric testing, the linguistic competence of most chatbots is mostly equivalent to that of humans, even if the extent of understanding is unclear \citep{mahowald2024dissociating}. Regarding types of intelligence beyond linguistic and logical, we return to this issue when we discuss embodiment. 

\section{Animal Intelligence}

Humans are not the fastest, toughest, most enduring etc. creatures, but we are distinguished for our cultural, technological, and scientific accomplishments -- our achievement in these respects surpasses that of all other creatures. This has allowed us to consider ourselves the most intelligent species. 
Even if this is true in some sense, there is a question of whether other species are intelligent like us, but to a lesser extent, less intelligent in fundamental ways, or perhaps in some ways more intelligent than humans have realised.

There are signs of intelligence throughout the animal kingdom (Bukart et al., 2017). For example, dogs have highly developed social awareness, including sensitivity to human body language, human gestures, and parts of language 
\citep{hare2002domestication,kaminski2004word}. 
Dogs are able to solve a variety of logical/ reasoning problems, such as a means end task \citep{range2011dogs} or spatial puzzles \citep{barrera2015effects}. 
Theory of mind is the ability to attribute mental states (such as beliefs and desires) to others. It is typically considered an integral part of intelligence, especially social intelligence \citep{navarro2022what}. 
There have been arguments that dogs and other primates have a theory of mind \citep{catala2017dogs,premack1978chimpanzee}, though there is controversy \citep{heyes1998theory}. 
Chimpanzees have extensive tool-using abilities, including for food gathering/ hunting, grooming, and communication 
\citep{koops2015chimpanzees}. Chimpanzees can recognize abstract symbols \citep{rumbaugh2003intelligence} and can innovate solutions to unfamiliar problems \citep{bandini2020innovation}. Similar evidence has accumulated for other species, e.g., corvids, who appear to have problem-solving abilities comparable to those of chimpanzees \citep{gunturkün2023birds}. Perhaps unsurprisingly, there is much evidence regarding musical intelligence in birds, e.g., relating to rhythm perception and processing \citep{fitch2015four,patel2006musical} or the structural properties of birdsong \citep{kawaji2024goal,zhang2023analogies}. Further, there is some evidence of musical synchronisation amongst birds, though for duetting birds at most \citep{hall2009review,mann2006antiphonal}. Overall, bird musical ability does not appear to match the complexity of human music \citep{lerdahl1983generative,levitin2006brain}. 

What about, for example, the homing instinct of a pigeon or the way a leopard calculates the exact time and distance before it pounces onto prey? For example, pigeons have been known to successfully return home across distances exceeding 100km \citep{schmidt1972homing}.
We would not call an individual pigeon intelligent for demonstrating its homing instinct, in the same way we would not call an individual human intelligent for learning to walk or ride a bicycle --- such skills are considered default competencies that most members of a species are endowed with, as a result of evolutionary adaptation. There is intelligence at the species level, allowing for different competencies, developed through evolution (e.g., \citep{jerison1988intelligence}). A similar case can be made regarding the remarkable ingenuity in, for example, the way ants or bees build their nests \citep{winston1987biology}. Note, linguistic competence could also be considered an instinct (e.g., \citet{pinker2015language}). It is not the linguistic competence as such which is the crucial point, we think, but rather the fact that complex thought is typically (not always) communicated linguistically. Modern chatbots are clearly linguistically competent. Are they also capable of complex thought to the extent that we can consider them intelligent? 

Self-interest is evident throughout animal behavior (for example, taking steps to avoid being eaten).
Self-awareness has also been studied in animals, including the famous finding of \citet{gallup1970chimpanzees} that chimpanzees, but not monkeys, learn to recognize a reflection in a mirror as their own.
It is tempting to assume that humans are uniquely self-aware, hence more intelligent than other animals, but attempts to describe or demonstrate something distinctly self-aware about humans are empirically challenging \citep{carruthers2009how}. Whatever form mental states may take, humans can communicate limited parts of these experiences to one another, yet lack any comparable window on the mental life of animals; hence, we should not take absence of evidence as evidence of absence.

Symbol processing has long been emphasized as a traditional core of intelligence \citep[e.g.][]{whitehead1927symbolism}. The recent ability of chatbots to manipulate linguistic symbols so fluently is disconcerting partly because, until now, this capacity seemed unique to humans. Yet, as \citet{torresmartinez2024embodied} argues, non-human animals possess distinct semiotic systems that shape their own forms of cognition. He suggests that robot and AI design should attend to these alternative systems to avoid reproducing the specifically human constraints embedded in our symbolic models of thought.

Beyond evolutionary intelligence, for some species, there is collective intelligence, the ability to achieve more than what an individual could accomplish, through cooperation. 
In the animal kingdom, for example, ants use pheromones to communicate about food sources or dangers and often work together in foraging tasks \citep{holldobler1990ants}; the way African wild dogs coordinate in hunting has sophisticated characteristics, including role differentiation and vocal communication (\citet{creel2002african};  for an example regarding see rats \citet{schuster2002cooperative}). 
However, no other species has been more successful in transcending what is possible for an individual than humans. \cite{fernbach2018knowledge} coined the term `community of knowledge', for the idea that even seemingly simple aspects of our lives are underwritten by massive knowledge contributions of which we are mostly ignorant. Our ability to form sophisticated communities of knowledge is fundamentally linked to the unique expressivity of human language (e.g., \citet{lakoff2003metaphors,pinker1995language}) 
and the fact that our language has a written form \citep{latour1992laboratory}. 
Written language has enabled global communities of knowledge; arguably,  it is the cornerstone of our civilization (\citet{goody1977domestication}; concerning the interaction between climate, the durability of early writing mediums, such as papyrus, and the emergence of early civilizations see \citet{gaudet2014papyrus}). 

Is there any evidence for art, science, technology amongst non-humans? Regarding art, some animals can learn to discriminate based on the aesthetics of pictures and show preference towards higher aesthetics (e.g., songbirds,  for musical stimuli; \citet{watanabe2012animal}). Some birds create and decorate elaborate structures to attract mates \citep{borgia1985bower} and some animals can create pictorial patterns in the sand or using rudimentary tools, which resembles artistic expression \citep{pele2021orangutan}. However, such evidence is a far cry from demonstrations that non-human animals show human-like appreciation of these behaviours and their outputs, in the way we do for art \citep{english2014painting}.  Regarding science, the scientific endeavour implies experimentation and exploration of testable hypotheses, to identify general principles and so generalize from known situations to novel ones. There is no evidence that animals can do science. However, there is some evidence that animals can experiment and adjust strategies, create artifacts, and develop rudimentary tools. For example, chimpanzees make tools to help them forage for ants or termites, through a process of successive improvement before the desired form is achieved, sometimes before actual tool use (indicating that they have an idea of the required form for a tool prior to use; \citet{boesch1990tool}); and New Caledonian crows have been reported to construct hooked tools, to aid them in insect capture \citep{hunt1973individual}. Moreover, there is evidence of animals adapting behaviour from trial and error and changing circumstances (e.g., \citet{fiorito1992observational}).

Such evidence indicates that animals are capable of technology and innovation, though it appears that their only path is through gradual trial and error, as opposed to the sudden realization which often accompanies human technological innovation (famous examples are penicillin and the discovery of the Benzene chemical structure, \citep{fleming1929antibacterial,kekule1865constitution}, as cited in \citet{rocke2015daydream}).  

\section{The Word and Western Culture}
\label{ref:western_word}


Electronic computing and artificial intelligence are recent products, driven by relationships between technology and science, that arose along with industrialization, in a culture that particularly prized written forms of intelligence.
This section summarizes aspects of these cultural developments, and how they influenced some of the development and attitudes towards artificial intelligence.

In standard historic accounts, western civilization is the product of Hebrew religion and Greek 
science, becoming the Judeo-Christian tradition. Both of these roots embody the distinctiveness of 
human origins and intelligence, with an emphasis on knowledge expressed in words. In Genesis (Ch 
3), Adam and Eve eat the fruit of ``the tree of the knowledge of good and evil'' and become self-aware. In Aristotle's \textit{Politics} (Bk VII Ch 3), ``Virtue arises through nature (\textit{physis}), habit (\textit{ethos}), and 
reason (\textit{logos})'', and while some animals demonstrate ethos as well as physis, only humans have 
logos. 
This combination became central to Christianity, whose most mystical Gospel starts with  ``In the beginning was the Word'' (\selectlanguage{greek}Ἐν ἀρχῇ ἦν ὁ λόγος\selectlanguage{english}, John Ch 1). 

This did not originally imply that knowledge and intelligence are expressed in \emph{written} words.
Many great scholarly traditions have emphasized oral learning, and many scriptural and epic works were known and taught long before they were written down in canonical manuscripts.  
Plato's \emph{Phaedrus} dialogue raises fears that those who rely on writing ``will not use their memories \ldots they will appear to be omniscient and will generally know nothing'' \cite{plato1953phaedrus}, analogous to arguments today that artificial language technologies reduce human skills.

\begin{figure}
    \centering
    \includegraphics[width=0.5\linewidth]{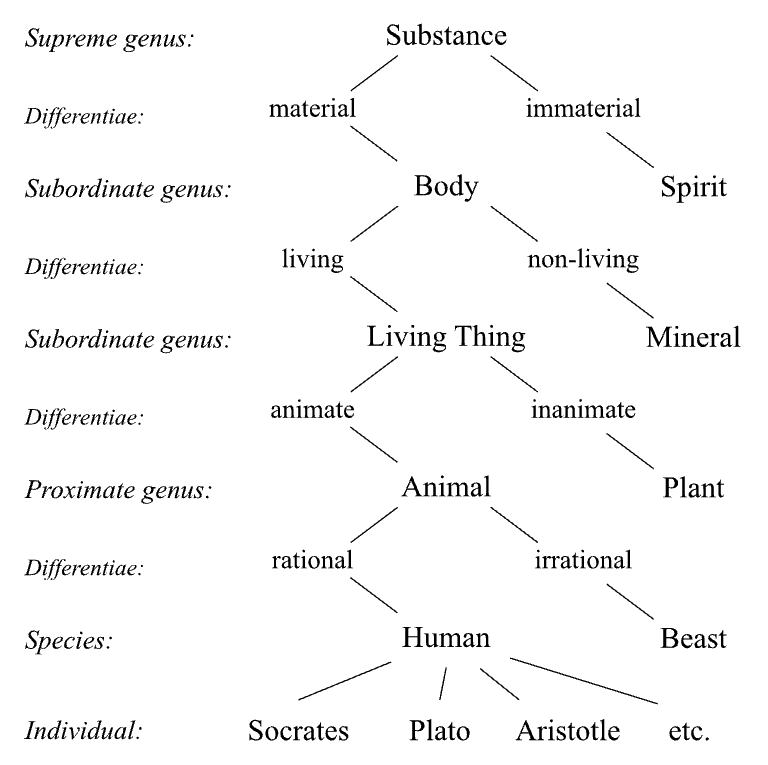}
    \caption{Tree of Porphyry, used to teach introductory logic 
    from 3rd to 19th centuries in European schools.}
    \label{fig:porphyry-tree}
\end{figure}

As Christianity became established, so did the belief that humans are verbal, rational animals, 
and that this is the divine core of our identity. 
Beyond religious education, this was taught in introductory logic and classification (we use classification in the logical sense) from the third century onwards,
using the tree of Porphyry (Figure \ref{fig:porphyry-tree}),
perhaps history's most famous decision-tree. 
It was used until the mid-1800s, when Boolean propositional calculus and truth-tables supplanted 
Aristotelean classification and syllogism in the teaching of logic \citep{boole1854laws}.
The claim that humans are uniquely rational was not just part of biology and theology, it was taught as a foundational principle in western logic!

The primacy of \emph{written} words was established gradually, and not only in Europe. The Qur’an emphasizes the sacred status of its own words and respect for Jewish and Christian scriptures, whose followers are described as ``people of the Book'' (Surah 5:65–69). In China, the tradition of Confucian literary scholarship became central to the famous civil service examinations, and by the Ming period (1368–1644) had become the main route to official status. In theory, this was a great source of social mobility, and historically the first great example of `standardized testing' \citep{elman2013civil}.

North-western Europe came to rely especially on writing as a source of truth and intelligence.
The widespread use of printing, and the Protestant Reformation, emphasized the view that the Bible itself was literally the Word of God, rather than the more mystical `Word made Flesh'.
In the Renaissance period, famous scholars were often multi-disciplinary and their works included sculpture, art, poetry and music; but for scholars of the later Enlightenment period, by `collected works', we tend to mean a book with just writings.
In music, a Western European tradition developed in which composers are celebrated, but by their `works' we mean written manuscripts. (In Western classical ensembles, the paper with the written notation is even called \emph{the music}, whereas in jazz ensembles, the paper documents are called \emph{charts}.)

The independence of science itself from roots in natural philosophy and common engineering practices also evolved during this era.
In the 1600s, the slide rule (based on Napier's logarithms) and the pendulum clock (based on Galileo's mechanics) were the first examples of theoretical discoveries, described in writing, leading directly to the creation of new devices.

Steam engines became pivotal in the industrial revolution and illustrate the emergence of a new, causal relationship
between technology and science. Ancient inventors, including Heron of Alexandria, had built demonstration steam engines, but they were never used to replace human labour. 
From the early 1700s, Newcomen's steam engines were in active use pumping water out of mines, 
but the steam was heated and then condensed in the same chamber, an expensive and dangerous
combination. Watt's brilliant innovation during the 1760s was to separate these duties between two different chambers, leading to reliable rotational power that could drain mines, then power a whole weaving factory, and eventually leading to steamships and locomotives. 
In trying to understand the relationship between heat inside boilers and mechanical work in piston chambers, scientists including Carnot, Kelvin, and Clausius developed modern thermodynamics between 1820s and 1860s. Thermodynamics was arguably the first branch of physics
that was deliberately motivated by the goal of optimizing an existing technology.

Maxwell's electromagnetic theory in the 1860s went leaps further, explaining the 
propagation of light waves in traditional optics and predicting the behaviour of an entire spectrum
of oscillations, invisible to human senses. During the 1880s, Hertz used these theoretical predictions
to transmit and receive real radio waves. 
This formed a new relationship: science had brought about a technology that was
beyond our natural imagination \cite[Ch 1]{oerter2006theory}.
In this new relationship, written works of science led directly to new kinds of machines,
and scientific progress enabled a landslide of innovations throughout the late 1800s and early 1900s.

Since that time, it has often been assumed that written science is the main source of technological inspiration and reliability.
Electronic computing is a clear example of technology that arose as the brainchild of written science,
and has inherited some of these presuppositions (with many very successful results).
Nonetheless, a small sample of inventions (fermentation, wheels, metalworking, textiles, clocks, gunpowder, printing, telescopes, steam engines, dried noodles, vacuum cleaners, velcro) demonstrate immediately that written science is just one path to invention \citep{mokyr1992lever}.

Anglo-American educational practices have particularly emphasized written assessments.
While the French \textit{Baccalaureate} and German 
\textit{Abitur} exams, formalized since the early-1800s,
included practical and oral components, British qualifications, like the Chinese civil exams long before, were standardized entirely by writing.
Given the emphasis on STEM education today, it is easy to assume that science education was developed along with science itself, but this would
be an anachronism: Newton, Priestley, Cavendish, and Darwin did not go to secondary schools with classes in physics, chemistry, and biology,
but instead, aside from geometry and algebra, a principal focus would have been classical languages and literature.
British ``grammar schools'' had existed since the late middle-ages and the redirection of endowment funds for ``other useful Branches of Literature and Science in addition to or (subject to the Provisions hereinafter contained) 
in lieu of the Greek and Latin Languages'' was controversial enough to motivate the Grammar Schools Act of 1840 \citep{grammarSchoolsAct1840}.
The use of written examinations as the measure of intelligence
was explicitly promoted in the Northcote and Trevelyan report
that led to the British Civil Service exams. Their main
stated goals included promoting merit over patronage, and also
``establishing a proper distinction between mechanical and intellectual labour'' \citep{northcote1854report}. 
While moral and physical fitness was to be 
vouched for in-person, for the purposes of choosing civil servants,
intellectual labour was identified with writing. Overall, such tests prioritized written over manual skills: to fix a broken boiler, the person who could 
write down the problem, order the replacement parts etc., was considered intelligent, whereas the person who actually mended the boiler less so. 

Well before the time the term AI was coined, assumptions about the relationship
between science, schooling, written words, and intelligence were part of the fabric of western European culture. 
For centuries, there were serious philosophical and scientific claims that abstract reasoning is impossible without language, and only 
recently, brain imaging studies have shown that these activities are
dissociated from one another \citep{fedorenko2024language}.

Language and reasoning abilities have clearly helped to distinguish us from other species, and writing as a technology has worked as a great catalyst for improving, and even helping invent, other technologies.
Not surprisingly, the most successful industrial societies have prized and cultivated writing skills.
However, such societies are prone to find the idea of a `thinking machine' threatening.
Turing even anticipated such cultural resistance, saying
``It is likely to be quite strong
in intellectual people, since they value the power of thinking more highly than others \citep[\S6.II]{turing1950computing}''.

\section{Artificial Intelligence Joins the Stage}
\label{sec:ai_intro}



The first electronic computers are a great example of a scientific discovery, based on written work, leading directly to new technologies.
The theory of computing was developed by Church and Turing in the 1930s, and Turing's calculating machines were built in the hidden limelight of wartime codebreaking in the 1940s.
By the end of the decade, Turing, von Neumann, and many collaborators had developed 
reprogrammable machines, whose instructions and calculations were stored in electronic
memory and were capable of performing thousands arithmetic calculations within seconds,
using cascading patterns of Boolean logic to process numbers represented in binary digits.

Turing's paper on \textit{Computing Machinery and Intelligence} in 1950 \citep{turing1950computing} is a foundation
of the field that was eventually called AI. Turing's proposal was particularly important for text processing, introducing a challenge called
the \textit{Imitation Game}. Two humans, or a human and a machine, try to imitate one another, communicating directly 
with a judge, with interactions restricted to the inputs and outputs of a teleprinter. 
Whether the judge 
can distinguish a machine from a human through this interface alone has come to be called the 
\textit{Turing Test}. 

Using the same distinction between physical and intellectual capacities as \citep{northcote1854report},
Turing reckoned that 
``The new problem has the advantage of drawing a fairly sharp line between the physical
and the intellectual capacities of a man.''
Turing carefully avoided the question of whether such a machine is
`thinking', saying instead that such behaviour would typically be \textit{regarded} as demonstrating intelligence. His paper examined common objections to the idea that such a machine
could be called intelligent 
(e.g., not having a soul or just learning `parrot-fashion'), arguing that they are rooted in prejudice and other mistakes,
and speculated that the most effective way to \textit{build} such a machine would be to
create a `child machine' with the programming and resources to be able to learn, and provide this machine with training experiences. 
He estimated that such a machine could be built by the year 2000 and, while that timeline was not achieved, Turing's paper broadly anticipates the current approach to language processing,
and some of the likely objections to ascribing intelligence to such machines.

Some of the core components of today's chatbots were pioneered in the 1950s and 1960s. Most famously,
Rosenblatt's invention of the perceptron \citep{rosenblatt1958perceptron} and Widrow and Hoff's use 
of continuous, gradient-based iterations to train perceptron networks \citep{widrow1960adaptive} are
generally considered the precursors of today's neural networks \citep[Ch 2]{raschka2022machine}. 
Representing words as vectors learned from data, usually called 
`word embeddings' since the success of the \textit{word3vec} package in 2013 \citet{mikolov2013efficient},
was pioneered in the SMART (System for the Mechanical Analysis and Retrieval of Text) project,
led by Gerard Salton during the 1960s \citep{salton1971smart}. 
The success of the SMART 
system was demonstrated using early statistical evaluations of precision and recall, developed
during the Cranfield experiments, whose purpose was indexing and retrieval techniques, particularly for scientific reference and military intelligence gathering \citep{cleverdon1966factors}. The Cranfield project was foundational in information retrieval, not only 
for its results, but also for establishing the practice of evaluating search engines by measuring the precision and recall of sets of search results, when compared with held-out test data. This practice was followed in the 1990s
in the Text Retrieval Evaluation Conference series \citep{harman1993first}, and submitting results to a ``TREC Style Evaluation'' is a key forerunner of many machine learning metrics and leaderboards, such as Kaggle and the MLCommons, used to compare models today.  

One of the key surprises from the Cranfield experiments was that automated indexing strategies, based purely on the words occurring in each document, outperformed indexing by human experts. This surprised, and initially 
dismayed researchers: \textit{``the results which seem to offend against every
canon on which we were trained as librarians''} \citep{cleverdon1966factors}. `Computer' had been a job title
for skilled arithmeticians during World War II, but since the rules of arithmetic had been formalized,
it stood to reason that a faster electronic machine that performed arithmetic could produce results that were
just as good. `Indexer', however, was a different matter: the job had been
performed by expert librarians since the 1800s, and the notion that skilled work with words could be done by a machine
just as well, if not better, was a shock. 


Given such strong foundations, it may be surprising that statistical language processing using neural networks
took so long to fully flourish. One reason is that it required decades of hardware advancements, including the
development of graphics processing units and their adaptation to neural networks, to realize the potential of neural
network methods. Another factor is that most research in AI and computational linguistics during the later decades of the 
1900s was focused elsewhere, particularly on symbolic, logical, and syntactic methods. In linguistics,  
Chomsky wielded great influence and his early work on syntactic structures claimed both that \textit{``The most 
that can be reasonably be expected of linguistic theory is that it shall provide an evaluation procedure for grammars''}
and \textit{``probabilistic models give no particular insight into some of the basic problems of syntactic structure''}
\citep{chomsky1957syntactic}. 
In AI, the book \textit{Perceptrons: An Introduction to Computational Geometry} \citep{minsky2017perceptrons},
 demonstrated key problems encountered by neural networks with the learning of non-additive functions, including
cases as simple as the logical XOR. As key inventors of early neural networks, Minsky and Papert emphasized 
in their book that their objections were not meant to divide the field, but rather that research on neural networks would fade
as other ideas took over; the 1988 edition included an epilogue saying
``\textit{In any case, the 1970s became the golden age of a new field of research into the representation of knowledge.}''
The new techniques listed as part of this `golden age' included word-expert parsers, nonmonotonic logic, and planning procedures,
which led many researchers to ``\textit{set aside their interest in the study of learning in favor of examining the representation of knowledge}.'' With hindsight, this is seen more as a dead-end than a golden age and the beginning of
a period that is sometimes called the ``AI winter''. (Ironically, the term AI winter was also coined by Minsky,
during the 1980s, as a warning against what might happen if expert systems did not bear fruit.)
But for some decades, symbolic logic was considered to be the core of artificial intelligence and linguistic theory \citep{jones1994natural} and the most established textbooks on mathematical and computational linguistics from
this era focus on discrete symbolisms, barely mentioning statistics, probability, or continuous mathematical 
foundations such as vectors and gradients \citep{partee1993mathematical,allen1995natural}.

Today's chatbots come from the very techniques that were supposed to have been left behind:
machine learning in general and neural networks in particular are the cornerstones of AI today.
During the 1980s, researchers including Hinton and LeCun developed the backpropagation method,
which enables deep neural networks to be trained efficiently, by passing feedback on classification results back
through layers in the network, using the chain rule from multivariable calculus \citep{rumelhart1986learning}.
Neural nets built in this way started to outperform expert systems at several tasks, for example image recognition for handwritten digits \citep{lecun1995convolutional}. In 1997, Deep Blue, a chess computer program invented by Hsu, a doctoral student at Carnegie Mellon University, defeated the world chess champion Gary Kasparov, relying on a massively parallel expert fuzzy-rule-based system and 
effective use of a dataset of Grandmaster games \citep{campbell2002deep}. In 2000s, the first neural net model for next-word prediction
was trained successfully to produce relatively consistent natural language phrases, over longer context distances
than those achieved with $n$-gram models. AI was beginning to produce systems that outperformed humans on complex symbolic 
tasks and, rather than doing this using rules created by human domain experts, this was achieved by leveraging fast massive
computation. 
The game of Go, more combinatorically demanding than chess by a wide margin, was mastered by the Alpha Go system,
using databases of real games for training, and then even more strikingly, by Alpha Go Zero, which learned 
solely by playing against itself, with no human supervision except for being given the rules and the resources to simulate play
\citep{silver2017mastering}.

Early on, attempts at AI development focused on creating systems which would excel at specific tasks, assuming that such systems put together would create a human-like, general intelligence: for example, Shannon
famously proposed that the game of chess would be an ideal challenge for developing intelligent systems, since it is 
\textit{generally considered to require ``thinking'' for skilful play} \citep{shannon1950chess}. 
Based on the combinatoric explosion of possible chess positions, Shannon argued that a brute-force approach, leveraging large computational power, would at best enable the creation of a weak chess player.
But, fast forward to a few decades later and it turns out that one can design an extremely effective chess system, mainly by applying brute-force computation over large datasets.
This is both an insight and a challenge to humans; 
some of the most poignant (albeit wry) reflections on this theme have been written by Kasparov himself, the first human world chess champion to lose to a computer (or as \citep{kasparov2017deep} reminds us, the last human to win a match against the world computer chess champion, a title he is likely to retain while records last).

AI agents have become better by carefully leveraging larger example datasets and computational resources, a process which contrasts with the
way humans accomplish similar expertise, typically requiring years of devotion \citep{gladwell2008outliers}.
Understandably, this is often unwelcome news to human experts, and the theme 
\textit{that general methods that leverage computation are ultimately the most effective, and by a large margin} has even been dubbed ``The Bitter Lesson'' \citep{sutton2019bitter}. 
But, the two questions, are we trying to construct systems which behave intelligently and are we trying to re-create human intelligence, are not the same, though the recognition of this difference has taken some time 
\citep[Ch 1]{russell2022artificial}. 

The large language models (LLMs) that drive modern chatbots are a striking example of this engineering approach. They are
built using neural networks to predict sequences of word-embedding vectors, the most recent key ingredient being
the use of attention mechanisms, which enable sequential behaviour to be modelled using parallel computation \citep{vaswani2017attention}. Subsequent models, including GPT-3 \citep{brown2020language}, Llama \citep{touvron2023llama}, and DeepSeek \citep{liu2024deepseek}, are trained and optimized to whole new levels of performance, though the core components
of word embeddings, feed-forward neural networks, and attention-heads using projection (with the whole system trained to
optimize a loss-function throughout the network using backpropagation) are fundamentally those established in the
transformer model of \citet{vaswani2017attention}.

These models use none of the syntactic parsers, logical forms, or knowledge representation techniques that dominated
much of late-1900s AI. Instead, they work much as \citep{turing1950computing} suggested: a `child machine' is 
initialized with enormous computational resources and relatively few built-in procedures, just enough to learn from masses of training examples, after which it performs well ---
so well that its output, as exemplified by modern chatbots, produces language successfully enough that it would be considered
a clear sign of intelligence in many established tests.


%

\section{History of (Artificial) Intelligence Tests and Moving the Goal Posts}
\label{sec:intelligence_testing}


Some of the early approaches attempted to test intelligence in terms of sensory ability \citep{galton1883inquiries,jensen2011theory}, 
an idea which was criticized by Wilhelm Wundt, who favoured complex cognitive thought, such as problem solving \citep{blumenthal1975reappraisal,wundt1896lectures}. 
Wundt's ideas influenced the first standardised intelligence test, the Binet-Simon scale, and the idea of mental age, as a way to characterise developmental delays in intelligence (\cite{binet1907enfants}). 

The Binet-Simon scale inspired later intelligence tests, such as the Wechsler Adult Intelligence Scale (WAIS; \citep{wechsler1939measurement}) and, in its US reincarnation \citep{terman1916measurement}, the Army Alpha (verbal-based) and Beta (non-verbal, for e.g. non-English speakers) tests, originally used for screening US Army recruits during the first world war \citep{yerkes1921psychological}. 
This work also led to the term intelligence quotient (IQ), the ratio of mental to chronological age \citep{stern1912psychologischen}. 
At around that time, researchers started recognising how some of these tests are culturally biased, favouring those from Western, educated, industrialised backgrounds \citep{bond1934education}. 
Such concerns led to the development of ostensibly culturally fair and non-verbal tests, such as Raven's Progressive Matrices \citep{raven1938progressive} and the culture-fair intelligence test \citep{cattell1943measurement}, 
an effort which continues to this date \citep{nisbett2012intelligence}. 
An interesting assumption in these proposals is that verbal ability is largely due to education and so should not be part of intelligence testing. 
Today, successors to the Stanford-Binet test \citep{terman1916measurement} and the WAIS test are still widely used, in e.g. clinical diagnosis, education, and cognitive functioning evaluations.
Pertinently, there is evidence that the more recent chatbots (such as GPT-4) can solve a variety of problems used in IQ tests, including in mathematics, verbal reasoning, comprehension of complex ideas etc. \citep{bubeck2023sparks}, 
as well as perform highly in standard college applicant aptitude tests, such as the Stanford Achievement Test (SAT) and the Graduate Record Examination (GRE; \cite{openai2023gpt}). 
Strong performance has already been observed with the type of problems typical in Raven's Progressive Matrices test \citep{webb2023emergent} and the recent versions of the WAIS. In February 2025, Grok AI announced expert-level achievements
on advanced math (AIME), science (GPQA), and programming (LiveCodeBench) tests \citep{xai2025grok3beta}.  Where recent chatbots (such as GPT-4) do not perform as well, this can often be traced to technical limitations of such models (e.g., visual processing; \citet{hernandez-orallo2017measure}). 

Using tests designed for human examinees to benchmark AI systems has attracted recent attention, enabled by the sophistication of today's AI models. In previous decades, AI 
testing methodologies have typically focused on basic competencies, such as proportion of working transistors. Software tests have traditionally been carried out at different levels: unit testing
checks that individual functions perform their calculations correctly; integration testing checks that the parts, 
when assembled, demonstrate the expected overall system behaviours; and regression testing checks that an updated 
version or specific instance of a system does not unexpectedly fall behind established performance expectations.
None of these processes has been called `intelligence testing', though 
there are analogies with the approaches of \citep{binet1907enfants} and educational testing since:
for example, the 2 times table is such a simple test, it should be passed by all students and assumed to be 
known thereafter. 

Evaluation in machine learning has gradually adopted a more statistical approach, as outlined in Section \ref{sec:ai_intro}. Instead of asking human experts to engineer rules for how to classify objects as a basis for (initially) relevance judgments \citep{cleverdon1966factors} and (later on) AI, human experts were asked to judge what the results should be. These judgements or annotations are then used as training examples for assessing AI accuracy.
Examples include relevance judgments for document retrieval, what characters should be recognized in 
a handwriting sample, and whether an email
message is spam or junk mail \citep{cormack2008email}.  
The use of standardized measures on particular datasets raises important questions, including sampling bias and whether a good benchmark result
guarantees appropriate behaviour in real applications. Another challenge is that human judges do not always agree: 
measuring and studying inter-annotator agreement, or inter-rater reliability, is an important part of the 
methodology \citep{hallgren2012computing}, though once an annotated dataset is available, it is common for 
machine learning evaluations to ignore this issue, even referring to labels from human annotators as `ground truth'.
Notwithstanding these issues, the gradual adoption
of established datasets and quantitative evaluations is part of what has enabled AI to make consistent progress since the
1990s, including the use of BLEU scores to evaluate machine translation \citep{bahdanau2014neural,vaswani2017attention},
and the development of comprehensive sets of language processing benchmarks designed to cover a variety of competencies
\citep{wang2018glue}.

Evaluation of systems by competing directly with human performance has been less of a motivating force in AI until recently. Initially, it was obvious that computers were faster and more reliable at arithmetic calculations
(this is what they were built for), so there was little point in comparing human with mechanical performance, 
just as we would not race a powered vehicle against a human runner. By contrast, most other human behaviours 
regarded as intelligent were clearly beyond early computers. Games like chess and Go
became exceptions, but organized competitions for Turing-like tests (particularly the prize offering by
AI inventor Hugh Loebner) have declined since the late 2010s \citep{singh2022survey} (though just 
as this paper was being written, the experiments of \citep{jones2025large} found that GPT-4.5 fooled human judges
into choosing it as the `real' human in an astonishing 73\% of cases).
The recent strong performance of AI systems on traditional intelligence tests for humans is not the culmination
of decades in pursuit of this goal: it is more the result of decades of progress in lower-level benchmark 
evaluations, with the surprising outcome that systems trained to predict next words in a document or appropriate
answers to relatively simple questions ended up performing well on tests designed for humans.

By the early 2010s, it was already clear that natural language processing systems could reach human-level performance on some
language tasks, including syntactic tagging, information extraction, and textual inference.
This motivated the development of new tests designed to highlight areas that are easy for humans,
and where AI was demonstrably still weak. For example, the Winograd schema challenge (named after a pioneering advocate for the study of `common sense' in AI) asked simple 
questions like:

\begin{quote}
{\textit{The trophy doesn’t fit in the brown suitcase because it’s too big. What is too big?\\
Answer 0: the trophy. Answer 1: the suitcase.}} \citep{levesque2012winograd}
\end{quote}

As emphasized by \citep{chollet2019measure}, a pioneer of more general tests in AI, challenges 
like this are designed to stimulate, not to disparage, progress in AI, and sure enough, several
systems had successfully tackled the Winograd schema challenge by 2020 \citep{kocijan2023defeat}.
\citet{chollet2019measure} proposed a set of challenges called the Abstract Reasoning Corpus (ARC),
which focus on shapes in grid patterns and how they might interact physically: for example,
it is easy for humans to see that a red shape is moved until it collides with a blue shape and to apply the
same `rule' immediately to the next scene. As of early 2025, performance of AI systems on ARC tests
has leaped dramatically and, while there is discussion on whether the test has been conclusively passed (due particularly to energy considerations -- the official ARC test included energy-consumption limits that were breached by GPT-03),
\citet{chollet2024o3} praises the groundbreaking progress in AI.

\begin{figure}
    \centering
    \renewcommand{\arraystretch}{1.5} 
    \begin{tabular}[t]{p{0.4\linewidth} p{0.4\linewidth}}
    \textbf{Prompt:} Please draw a diagram of a chessboard at the beginning of the game.
         &  
    \textbf{Prompt:} Please draw a picture of a C major scale ascending from middle C in musical notation. \\
    \bf{ChatGPT (recent)} \\
    \includegraphics[width=\linewidth]{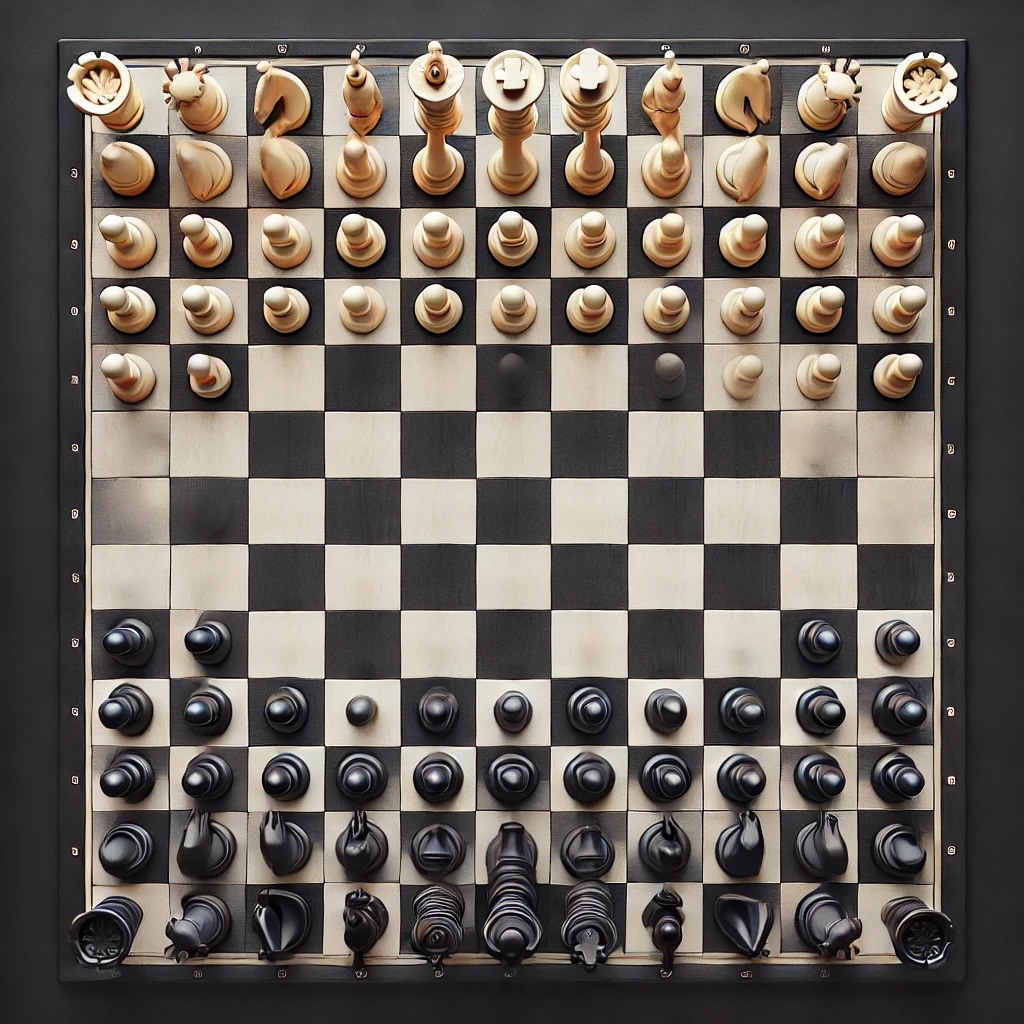} & 
    \includegraphics[width=\linewidth]{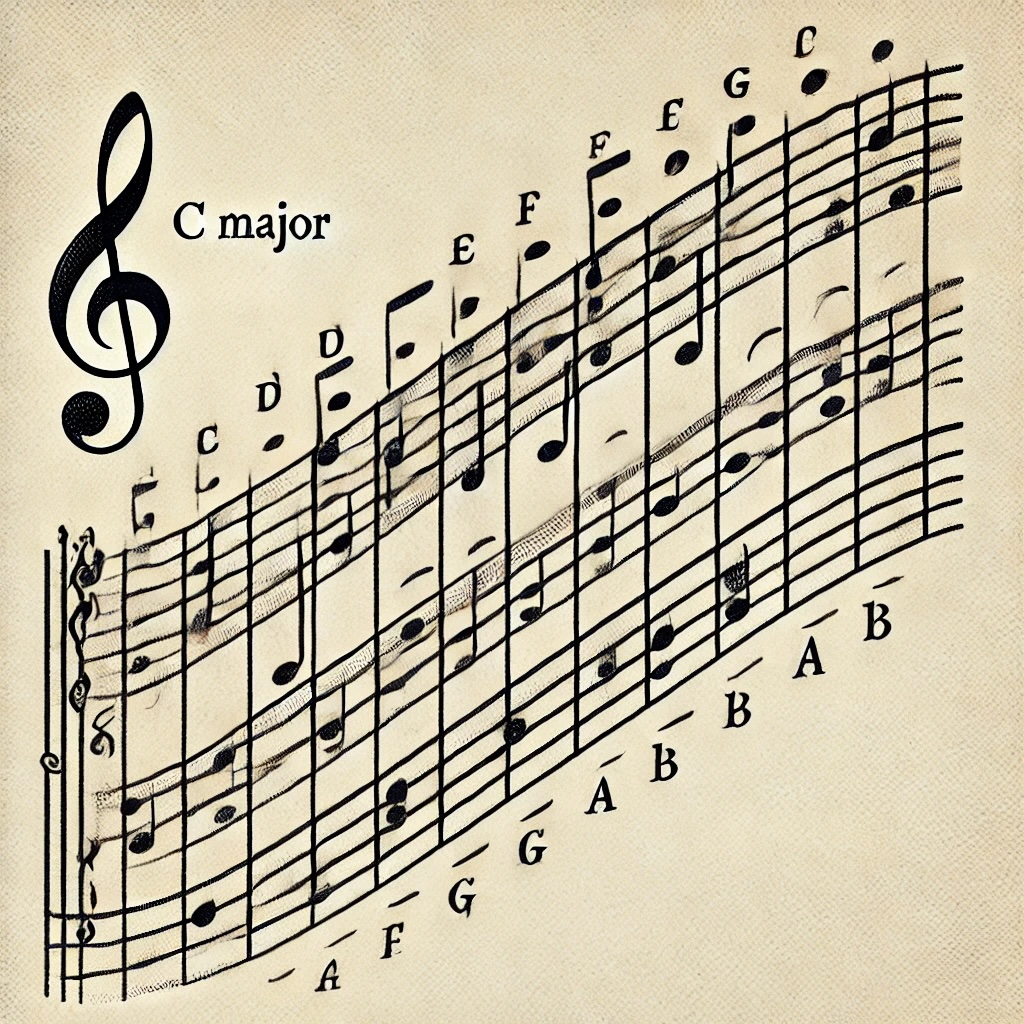} \\
    \bf{ChatGPT 4o (current)} \\
    \includegraphics[width=\linewidth]{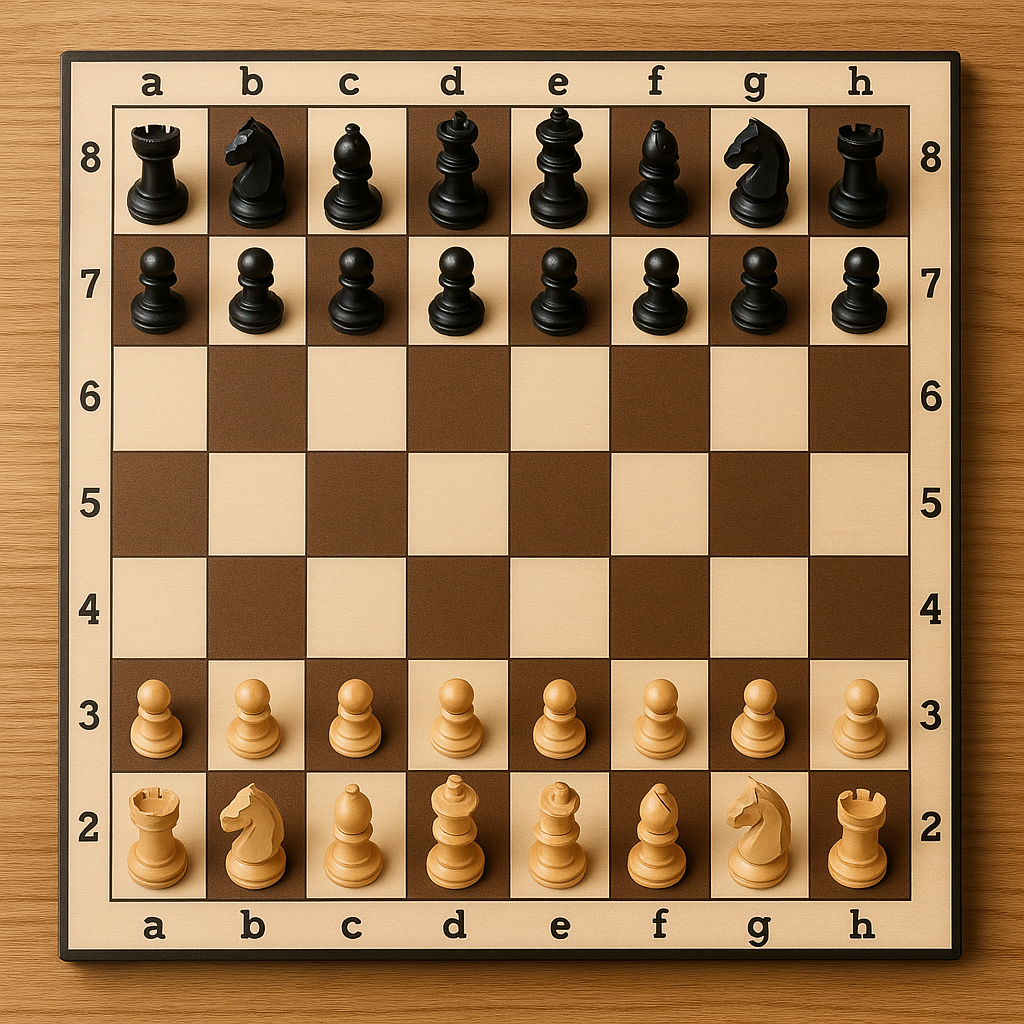} & 
    \includegraphics[width=\linewidth]{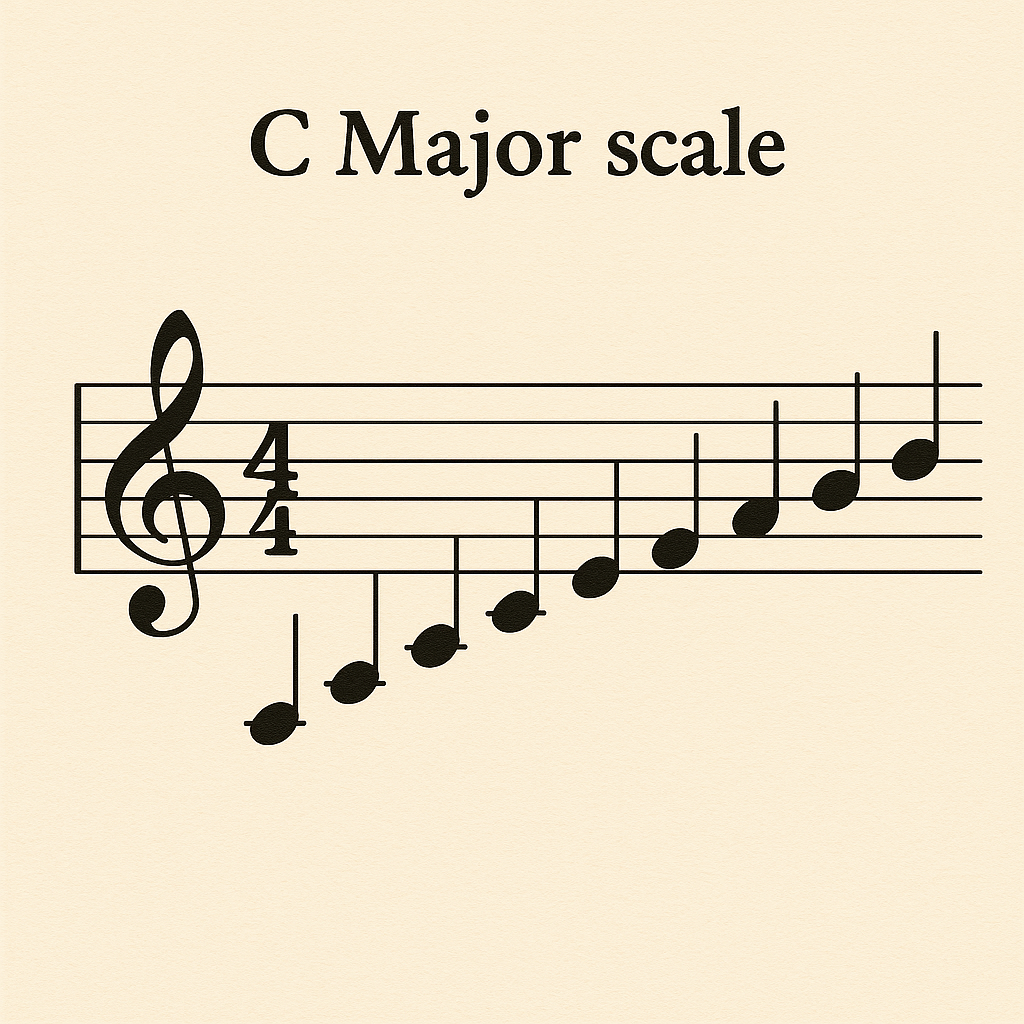} \\
    \end{tabular}
    \caption{Examples of image generation, showing artistic flair but lacking accuracy. Images generated by using ChatGPT, 2025-03-01 (above), and 2025-04-07 (below, after 4o upgrade).}
    \label{fig:chess_scale_pics}
\end{figure}

It is easy to find evidence that connecting words with images in an exact fashion is still challenging for
AI systems. State-of-the-art AI systems can generate stylish images
from text prompts, but without preserving key facts: for example, a C major scale should have 
the notes CDEFGABC, and a chessboard at the beginning of the game should have exactly 64 squares and 32 pieces;
AI-generated images still fail to meet these simple requirements, even though they can be very creative (Figure \ref{fig:chess_scale_pics}). Such gaps are perhaps even more surprising given the achievements of these models at tasks considered hard for humans (including the skilful drawing of shape,
texture, and shading). 
Though note,  more recent chatbot versions can produce outputs already better in diagrammatic accuracy.



Summarizing the apparent trends, in many IQ or aptitude tests, chatbots already outperform most humans. Such demonstrations invite a criticism that AI systems can accomplish this with vastly more resources, compared to human brains. For example, GPT-4 has approximately one trillion parameters and an estimated cost of about \$100M to train, while Deep Blue had about 14,000 specialized chess chips. 
However, comparing complexity between human and artificial agents is not straightforward. First, the human brain is thought to have more than 80 billion neurons and 100 trillion neuronal connections \citep{azevedo2009equal,herculano-houzel2012remarkable}, 
leading some scientists to claim that it is ``\ldots the most complicated object in the known universe''. 
Second, the brain has optimised many hardwired structures, making it efficient in cognitive tasks related to our physical constraints, environment, and needs. 
This optimisation has taken place across hundreds of millions of years for some generic abilities (e.g., visual perception; the time references the emergence of the first mammals, \citet{kemp2005origin}) and across hundreds of thousands of years for cognitive abilities more specific to humans 
(the time references the appearance of Homo Sapiens, \citet{hublin2017new}). 
So, while it might appear that some of these AI systems have frightening complexity, our brains are not exactly simple either, and use vastly more neurons that would be logically necessary to solve some problems.

Another possible objection is that some of the AI solutions to particular problems may appear uncompelling, because they involve brute force approaches. For example, Deep Blue can evaluate approximately 200 million positions per second and, even though it uses various pruning algorithms to reduce calculation, it still searches through millions of moves \cite{campbell2002deep}. Cholet's \citeyear{chollet2019measure} ARC test, intended to be the new gold standard regarding testing for AI, has recently been partially overcome, by GPT-o3 \cite{chollet2024o3}. Recent developments leading to this achievement include 
Chain-of-Thought prompting, which automates the process of breaking a problem into basic steps \citep{wei2022chain}, and test-time scaling, which devotes extra computational resources at inference time to check and revise answers \citep{muennighoff2025s1}. The techniques used by such systems could be described as successful implementations of the proverbial advice ``think before you speak''. Arguably, the success of inference-time improvements is just the application of brute force computation outside the training process.
However, the human brain is also an example of a brute force approach to intelligence, if not in how we solve problems, certainly in how evolution has optimised cognitive processes for the range of problems relevant to us. 

We suggest that the reluctance to call e.g. Deep Blue intelligent is less so about its brute force approach and more about domain specificity: we are unlikely to call intelligent an agent who excels (however exceptionally) in a single domain (cf. the distinction between skill and intelligence \citep{chollet2019measure}). 
We do not call a calculator intelligent, but a small child who could multiply 4-digit numbers in their head would be considered an intellectual prodigy.
A teenager who wins a spelling-bee competition is often described as `smart', and while any literate human could do just as well by using a dictionary, we expect that the spelling-bee champion is also good at remembering other things.
Estimating (and sometimes overestimating) the extent to which success on one test demonstrates aptitude for other activities has long been recognized as a challenge for test-driven assessments \citep{gardner2011frames}.

It may look that, as soon as an artificial system passes the current gold standard test for intelligence, we decide that the test was inadequate and invent a more elaborate one. That is, every time an AI test is passed, we move the goal posts and change our definition of intelligence, (apparently) deliberately so as to keep out the most sophisticated AI system. A critical perspective would be that we just rig the game to retain a perception of superiority. 
Some such examples are noted above (e.g., chess), but there are others: having a flawless conversation with another human or translating a document into other languages would have been unquestioned signs of intelligence, until computers could do such tasks. An alternative perspective is that revisions in AI tests simply reflect an increasing understanding of what intelligence is.  

Our belief that machines are likely to outpace humans in all clearly-definable tasks is partly because of human ingenuity and partly due to our lack of scruples (at least so far) when dealing with non-humans.
Pulling open neural networks and language models to find how best to improve them is always an option, which gives us options for improvement that we can explore systematically.
(This advantage is alluded to already in \citet{lecun1989optimal} with the paper title ``Optimal Brain Damage'' --- something we can explore in neural nets and not people.)

A possible limitation in considering chatbot capabilities in intelligence testing is a historic focus on the English language. For example, early language models tended to use English, as it has been in English that most data has been readily available. 
This has led to claims that AI agents are poor at interlingual communication. For example,
``AIs are usually programmed only in a single language'' \citep{nikolarea2024can}.
However, when such problems are recognized within the AI field, they often become superseded.
With appropriate training data, small multilingual embedding spaces have been built since the work of
\citet{widdows2002parallel}. Since the launch of BERT and mBERT \citep{devlin2018bert}, multilingual spaces have been standard \citep{pires2019multilingual}.
Returning to the point about 'dissection', nowadays, we can look inside many different embedding spaces in different layers, enabling machines
to cooperate with one another by exchanging non-linguistic tokens from latent spaces \citep{dessi2021interpretable}; one of the most pressing research challenges is to make such 
automatically-derived interlingua meaningful to humans \citep{karten2023interpretable}.
With AI models, we can explore, change, and experiment with every parameter in every layer,
and, while modern techniques such as fMRI scans have given some insight on activity in the human brain, there is no comparison with how transparently we can see and change each individual activity in AI models.

The ability to monitor different levels of activity in AI models also bears on questions of self-awareness.
The notion that an AI system cannot be self-aware like a human is one of the objections considered by \citet{turing1950computing},
who regards it as in fact an appeal to solipsism (\emph{self}-awareness is a personal experience, hence not something an individual can claim on behalf of a species).  
There is some work arguing that modern chatbots have an impressive range of metacognitive abilities (\citet{li2025aiawareness}; but see also \citet{nikolarea2022translators}). 

However, more generally, instead of asking whether chatbots can experience self-awareness, we can focus on their behaviours. When asked about a conversation, chatbots are increasingly able to explain why they responded in a particular way, and can sometimes recommend improvements in their own configuration to avoid problems recurring. (For example, they can suggest updates to {\small{\texttt{AGENTS.md}}} files, which are used by developers to control coding agents.) 
From an engineering point-of-view, it does not matter whether the agent experiences self-awareness: but if it can monitor and improve its own behaviour, it becomes self-aware in the way that matters to its collaborators.
As with other tasks and evaluations, if self-awareness is measured by answering introspective questions, AI will develop a foreseeable advantage, partly because we have more options for engineering the correct behaviour with chatbots than we do with ourselves.

\section{The Relation between Embodiment and Intelligence}
\label{sec:embodiment}

Human intelligence is tied to the way we perceive and manipulate objects in the world \citep{wilson2021embodied}), as well as our cultural and social context \citep{cobb2001situated}. One can ask what is the value of intelligence for solving embodiment problems --- we have developed a wide range of tools, which allow successful embodied problem-solving. 
Some researchers go as far as to suggest that all knowledge we have is either embodied or has its origins in embodied knowledge \citep{johnson2008metaphors,lakoff1999philosophy}.
For example, even in the most abstract scientific theories we have constructed, such as quantum mechanics, some of the basic ideas are extensions of embodied concepts (e.g., waves and particles); the most profound conceptual challenges in such theories come from ideas with no embodied analogues (e.g., superposition, non-locality; \citet{einstein1935can}).

Physical embodiment is certainly key to human intelligence (that is, our particular embodiment is tied to our notion  of intelligence) and many aspects of language reflect metaphorical extension of meaning from embodied concepts. For example, as soon as we say a problem is \emph{hard} and takes a \emph{long} time to solve, we demonstrate a spatial grounding to our concepts \cite[Ch 1]{torresmartinez2022metaphors,widdows2004geometry}.
Nonetheless, chatbots are adept at using these words: it is a great surprise that artificial agents talk as if they understand the world, even when their experience is reduced to digitized text \citep{summerfield2024strange}.

Could one rephrase the main question in this work from ``Are chatbots intelligent'', to ``can chatbots think''? 
Turing predicted that \emph{``at the end of the century
the use of words and general educated opinion will have altered so much that one will be
able to speak of machines thinking without expecting to be contradicted.'' \citet{turing1950computing}}
Turing's timeline estimate turns out to have been optimistic, but a quarter of the way through the next century, we do have models
that use various ``thinking modes'', and we are used to messages saying ``Thought for X seconds''.
The processes themselves are partly inspired by human approaches to problem solving, for example, by breaking reasoning down into steps and checking each step \cite{wei2022chain}.
Following technology history, we can expect that a chatbot \emph{thinking} something will be as commonplace as a book \emph{saying} something.
The objection that it is not \emph{really} thinking will remain active, and this concern remains important:
the situation is similar to Plato's concerns in the \emph{Phaedrus} dialogue, which urge the readers themselves to remember that \emph{reading} something is no substitute for truly \emph{knowing} something.

As Gardner’s account of musical intelligence reminds us, not all thinking takes linguistic form \citep{gardner2011frames}.
In \emph{Phaedrus}, Plato joins \selectlanguage{greek}λόγος\selectlanguage{english}  
with rhythm, memory, and harmony as qualities of a well-ordered soul; 
learning ``by heart'' and reasoning through sound or motion are ways of knowing, and Plato places them closer to the psyche 
(\selectlanguage{greek}ψυχή\selectlanguage{english}, often translated as soul) than the written word.
These attributes vitally involve embodiment: so the importance of embodiment is deeply established in Western philosophy, 
but has become de-emphasized.

This helps bring into focus the anthropocentric bias throughout the literature regarding human intelligence and AI tests. As we saw, this has been recognised, in that there has been effort in inventing `disembodied' tests of intelligence, not tied to our specific embodiment constraints. (We have also shown that this is not a universal approach, but one that has been dominant in industrialized Western societies since the 1800s.)
This approach has led to an over-emphasis on intelligence tests based on language and manipulation of information through linguistic means. 
Chatbots expose a fatal weakness to such tests, because, even though they are supremely capable wordsmiths, we are left uncertain as to whether they are intelligent or not and wondering if humans should be considered as genuinely inferior if chatbots beat us at reasoning challenges.
This tension reflects what \citet{torresmartinez2025dehumanizing} calls the dialectic of dehumanizing the human while humanizing the machine, a pattern that complicates how consciousness and agency are attributed across biological and artificial systems.

Arguably, as shown in Figure \ref{fig:chess_scale_pics}, key problems with state-of-the-art chatbots are less so about the question ``Can chatbots think?'' but rather about the question ``Can chatbots see what they are doing?'' --- that is, the problem is not about the thought process, but about basic awareness. 
Indeed, Figure \ref{fig:chess_scale_pics} highlights major problems with state-of-the-art chatbots, 
which translated into human terms would not be ``Can chatbots think?'' but ``Can multimodal AI systems
see what they're doing?'': the problem is not the quality of the thought process, but lack of perceptual ability.

This challenge has been well-recognized and addressed during 2025, by vision-aware models including Microsoft Omniparser \cite{lu2024omniparser}: it is already the case that state-of-the art chatbot tools for coding include interfaces for sharing images including screenshots. Limitations on the chatbots' contextual awareness remain, but are partly by-design, for example, the chatbot might be restricted from seeing menu options that elevate its own permissions.
Some vision capabilities are still in their infancy, but we are getting used to AI agents `seeing'. 
(We are already used to voice tools like Siri and Alexa `hearing'.) 

It follows that to understand human intelligence, we have to consider embodiment. Likewise, some artificial intelligence researchers have argued that artificial agents need some form of embodiment (e.g., through robotics; 
\citet{anderson2003embodied}), before they can develop intelligence.  Regarding chatbots, given that they lack any aspect of embodiment and their only basis for learning is input from language, it is unsurprising that any intelligence they develop would be different from ours. However, ideally our notion of intelligence would be versatile enough to recognise different kinds of intelligence (e.g., based on agents embodied in different ways or no embodiment at all). 
That is, to understand what it means to be human and human intelligence, we need embodiment, but our notion of intelligence in general should not be tied to a particular embodiment. Of course, historically we (humans) have been appropriating the notion of general intelligence to mean human intelligence, though we (the authors) think current evidence suggests otherwise (that is, that there can be general intelligence unlike ours). 

The discussion of embodiment can perhaps help disentangle skill from intelligence, a challenge which arises both for human and artificial intelligence (c.f., Aristotle's \textit{Metaphysics}, Bk 1, Ch1, and \citep{chollet2019measure}).
We are part of a culture that has assumed, for some centuries, that wordsmithing is intelligence, whereas manual dexterity is ``mere'' skill, and the use of physical force is sometimes disparaged as brutish. 
A cardinal example is the phrase ``the pen is mightier than the sword'' --- an adage celebrated even on the walls of the Library of Congress (the quote is from an 1839 play by Edward Bulwer-Lytton, in the mouth of Cardinal Richelieu, First Minister of France from 1622 to 1642). If we remove the warlike associations and propose more blandly that ``the pen is more useful than the knife'', then the claim is
obviously not true in general and depends on the work to be done.
For example, knives did not depend on pens for their invention, but the other way round. 

So, is the use of a pen or a knife more indicative of human intelligence?  
During the period since the 1800s, when `intelligence' has been defined and measured by success at different tasks, 
those tasks have tended to measure skills related to Gardner's linguistic and logico-mathematical intelligences \citep{gardner2006multiple}, often ignoring skills related to (say) bodily-kinesthetic or personal intelligences.
One way to address these questions is by arguing that intelligent behaviour does not depend on whether the task involves language or mathematics or manual labour etc. Rather, the key factor is the complexity of the task.
And, as noted by \citet{kasparov2017deep}, we have a long history of overestimating the complexity of intellectual tasks (such as remembering chess positions), based on the difficulty most of us experience with this, compared with the ease with which we can make a cup of tea.

\section{Concluding Comments}

Chatbots have fundamentally challenged our notion of intelligence. One reason is the role of (especially written) language in human intelligence and civilization. 
Language has been a key facilitator of progress in our species, by allowing inventions which persist beyond the individual, collaboration/ role differentiation, and communities of knowledge \citep{fernbach2018knowledge}. 
Historically, human linguistic ability has been pillar in our self-concept of intelligence and assumption of superiority --- so, with the emergence of chatbots, an observer might say that we have to either declare chatbots intelligent or argue that being a wordsmith is no longer enough, and change the rules for determining intelligence that we have insisted on for centuries. 

Given the prowess of chatbots on IQ tests and their many offshoots, it is tempting to look
for definitions of intelligence that emphasize other aspects of human achievement.
Even chatbots designed for coding often lack the ability to observe what's going on in different 
parts of a computer screen, and they can propose (and sometimes execute) syntactically-correct changes to code that have alarmingly bad consequences \citep{kim2025vibe}.
It is thus tempting to argue that AI still lack `general human-level intelligence',
on the grounds that some of the most advanced systems available clearly lack some of the basic physical and social intelligences observed in many animals.

However, while such arguments have been used to explain the historic success of humans in terms of `superior intelligence', they are likely to become obsolete due to progress in robotics and cooperative agents.
The potential for incorporating large language models in embodied agents is demonstrated by authors including \citet{zhang2023building}, and robots that dance and perform typical human physical activities are already being marketed for consumer use.
It is easy to anticipate machines that will win at tennis before long, in the same way that, by the early 1990s, it had become easy to anticipate that sooner or later machines would win at chess \citep{kasparov2017deep}.
The prototypical evolutionary example of human prowess in embodied and social intelligence might 
be reflected in the collaboration in a hunting or war party. However, it is hard to imagine that even the best human commandos would defeat a fleet of military drone hunters within a few years.
We may object to armed drones on moral grounds, but semi-autonomous armed drones and components are already commercially available \citep{summerfield2024strange}.

On the initial question, ``Are humans uniquely intelligent?'', chatbots demonstrate that humans are \emph{no longer} unique at demonstrating the symbol-processing intelligences that have been prized and measured since Victorian times. 
While this conclusion does not contradict any established scientific test, it is disruptive to the scientific establishment.

It is nonetheless obvious that chatbots alone do not demonstrate Artificial General Intelligence (AGI), in the sense of performing all cognitive tasks at least as well as a human --- in 2025, glaring common sense blunders are still far too prevalent.
For some years, training bigger and better chatbots led to such improvements that the approximation to human conversational intelligence is often surprisingly good \citep{summerfield2024strange}, but this trend has
already been superseded: advances in 2025 have highlighted reasoning models, and frontier tasks now include
semantically-accurate video generation. 

This widely accepted definition of AGI is explicitly anthropocentric. 
Should humans be accepted as a general standard of intelligence for all cognitive tasks?
Even by 1950, Turing observed that humans no longer had any hope of performing arithmetic as fast as machines: so if arithmetic is one of the cognitive tasks, humans cannot hope to surpass the `general intelligence' of machines.
Key advances in the scientific understanding of space and time have come from realising that man is \emph{not} the measure of all things, and human scales are small compared to the universe.
Truly general intelligence need not be limited by the intelligence of a human; we hope that allowing ourselves to distinguish these will eventually help us to understand intelligence \emph{and} humans better.

It increasingly appears that chatbots could replace humans in any task which is reduced to wordsmithing and the simple but somewhat embarrassing truth is that unique human competencies correspond to the animal stuff --- tasks in which we excel because of our particular embodiment, and for which a human empathic connection is useful.
However, given the rapid development of multimodal AI models and semi-autonomous robots, this assessment may  become dated within a few years.
Of course, replacing tasks does not always equate to replacing jobs: 
in some professions, including medicine and social work, reducing the burden of paperwork is a goal in itself \citep{johnson2021electronic}.
This is not necessarily a bad readjustment: perhaps if we, as a species, revised our perception of how unique our intelligence is, we might approach each other, our environment, and other species with a bit more humility.

\section*{Financial Support}

EMP was supported by AFOSR grant FA8655-23-1-7220

\section*{Competing Interest statement}

The authors declare no competing interests and no conflicts of interest.

\bibliographystyle{apalike_dense}
\bibliography{llms_bio}

\end{document}